\documentclass[10pt,twocolumn,letterpaper]{article}

\usepackage{cvpr}              

\usepackage{graphicx}
\usepackage{amsmath}
\usepackage{amssymb}
\usepackage{booktabs}
\usepackage[pagebackref,breaklinks,colorlinks]{hyperref}

\usepackage{algorithm}
\usepackage{algpseudocode}

\usepackage[capitalize]{cleveref}
\crefname{section}{Sec.}{Secs.}
\Crefname{section}{Section}{Sections}
\Crefname{table}{Table}{Tables}
\crefname{table}{Tab.}{Tabs.}

\def\confName{CVPR}
\def\confYear{2023}

\begin{document}

\title{Rewarded meta-pruning: Meta Learning Using Rewards for Channel Pruning}

\author{
    Athul Shibu, Abhishek Kumar, Heechul Jung, Dong-Gyu Lee\\
    Dept. of Artificial Intelligence, Kyungpook National University\\
    {\tt\small \{athulshibu, abhishek.ai, heechul, dglee\}@knu.ac.kr}
}
\maketitle

\begin{abstract}
   Convolutional Neural Networks (CNNs) have a large number of parameters and takes significantly large hardware resources to compute, so edge devices struggle to run high level networks. This paper proposes a novel method to reduce the parameters and FLOPs for computational efficiency in deep learning models. We introduce accuracy and efficiency coefficients to control the trade-off between the accuracy of the network and its computing efficiency. The proposed Rewarded meta-pruning algorithm trains a network to generate weights for a pruned model chosen based on the approximate parameters of the final model by controlling the interactions using a reward function. The reward function allows more control over the metrics of the final pruned model. Extensive experiments demonstrate superior performances of proposed method over the state-of-the-art methods in pruning ResNet-50, MobileNetV1 and MobileNetV2 networks. 
\end{abstract}

\section{Introduction}
\label{sec:intro}

Convolutional Neural Networks (CNNs) have been shown to achieve state-of-the-art results in various computer vision tasks \cite{karpathy2014large, huang2017densely, lee2022joint, lee2016human, lee2019prediction}. However, training the parameters of a CNN requires a significant amount of  labeled data. Furthermore, a large amount of hardware resources are also required for training a large amount of training data. Recently, network pruning has become an important topic to simplify and accelerate large CNNs \cite{huang2018learning, tian2019meta,lee2016human, yamamoto2018pcas}.

Many issues are at stake when trying to prune networks, such as structure \cite{li2016pruning}, continuity \cite{louizos2017learning} or scalability \cite{liu2018rethinking}. There are primarily two ways to compress neural networks: weight \cite{frankle2018lottery, su2020sanity, bouchard2009randomized} and channel pruning \cite{deng2019arcface, liu2021discrimination, elkerdawy2022fire}. Reducing parameters by pruning connections is the most intuitive way to prune a network. Weight pruning consists of identifying low-performing weights to be pruned \cite{han2015deep}. This involves simply removing weights with small magnitudes, which is easy to implement \cite{han2015learning}. 
However, most frameworks cannot accelerate sparse matrices during computation. In order to get real compression and speedup, it requires specifically defined software \cite{huang2018learning} or hardware \cite{han2016eie} to handle the sparsity. The actual cost is not impacted no matter how many weights are pruned. 

Therefore, channel pruning which involves removing whole filters instead of simply reducing the weight values to zero is preferred instead. \cite{kruschke1991benefits, liu2017learning}. 
This preference stems from the fact that channel pruning could remove the whole filters, creating a model with structured sparsity \cite{he2022filter}. With structured sparsity, the model can take full advantage of high-efficiency Basic Linear Algebra Subprogram (BLAS) libraries to achieve better acceleration. This makes the pruned model more structural and achieves practical acceleration \cite{han2016eie}.

MetaPruning \cite{liu2019metapruning} is one such channel pruning approach that can achieve the acceleration of CNNs. The central approach to MetaPruning is to generate weights for pruned structures instead of pruning weights or filters of the existing network. The accuracy of the untrained models is computed to rank each Network Encoding Vector (NEV). Evolutionary algorithms, which are motivated by processes of natural evolution \cite{kumar2019spherical}, are used to find the NEV that produces a model of the highest accuracy. Consequently, however, MetaPruning is only able to choose the best accuracy chosen from a set range of FLOPs. So the algorithm finds the highest accuracy within the predetermined range of FLOPs instead of trying to find the proper balance between sacrificing accuracy and reducing FLOPs.

This paper tries to address this issue. In the proposed Rewarded meta-pruning, instead of finding NEVs that produce the highest accuracy, the model tries to balance the accuracy with the FLOPs of the network to find the highest accuracy possible for the given FLOPs. In MetaPruning, the reward is directly proportional to the accuracy, because the reward is accuracy. So the increase in reward of subsequent mutations is lower compared to those of Rewarded meta-pruning, where the reward is directly proportional to the square of accuracy. At the same time, Rewarded meta-pruning is able to control the FLOPs of the final model by computing a score that takes both the accuracy and FLOPs into account to find models with high accuracy and low FLOPs. This score, which is the reward, can be further tweaked to include various parameters and control the metrics of the pruned model, as well as how the parameters interact with each other.

Our contribution lies in three folds:
\begin{itemize}
    \item We propose a channel pruning method, Rewarded meta-pruning, that can learn how to assign weights to pruned networks. 
    \item We explore the importance of reward functions and the characteristics to define an effective reward function.
    \item We experimentally show the superiority of the proposed pruning method on publicly available pre-trained CNNs; ResNet-50, MobileNetV1, and MobileNetV2.
\end{itemize}


\section{Related Work}
\label{sec:related_work}

The Lottery ticket hypothesis states that a randomly initialized dense neural network contains a subnetwork which, when trained in isolation, can yield results as well or even superior to the original network \cite{frankle2018lottery}. In other words, a standard pruned network can have the same, if not higher accuracy, than the original network. There are several methods to find the right tickets.

\textbf{Unstructured network pruning}: Various random pruning methods \cite{frankle2018lottery, su2020sanity, bouchard2009randomized} rely on pruning the parameters randomly based on various factors of the weights. \cite{ye2018rethinking} uses the L1 and L2 norm of each weight to compute their importance. The final pruned model is generated by pruning the less important weights. \cite{he2019filter} computes the geometric median of the weights while \cite{molchanov2019importance} uses the complex Taylor series expansion to calculate the weight function. Other weight pruning methods like \cite{luo2020neural} and \cite{liebenwein2019provable} use KL-divergence importance and Empirical sensitivity of the weights respectively to prune them.

\textbf{Structured network pruning}: In other approaches, AutoPruner \cite{luo2020autopruner} integrates filter selection into the model training so that the finetuned network can select unimportant filters automatically. Sparse Structure Selection (SSS) \cite{huang2018data} proposes the introduction of a new parameter, the scaling factor, which scales the output of specific structures. The sparsity regularization on these scaling factors pushes them to zero during training. Discriminative-aware channel pruning (DCP) \cite{NEURIPS2018_55a7cf9c} can find channels with true discriminative power and updates the model by pruning stage-wise using discrimination-aware losses. Adaptive DCP \cite{liu2021discrimination} introduces an additional discriminative-aware loss using the \textit{p}-th loss, and additional losses such as additive angular margin loss \cite{deng2019arcface}. AutoML for Model Compression (AMC) \cite{he2018amc} leverages reinforcement learning to automatically sample the design space and improve the model compression quality. Simpler methods like HRank \cite{lin2020hrank} determine the rank of feature maps generated by filters to rank the filters and their effectiveness on the final accuracy. This however takes more epochs to train after pruning. \cite{ye2020good} leverages the Lottery ticket hypothesis to greedily search through a network, finding subnetworks with lower loss than networks trained with gradient descent. Pruning algorithms that are inspired by Hebbian theory, like Fire Together Wire Together (FTWT) \cite{elkerdawy2022fire}, prune filters based on the binary mask of each layer and the activation of the previous layer.

\textbf{Meta-learning}: Meta-learning is the learning of algorithms from other learning algorithms \cite{vanschoren2018meta, finn2017model, liu2019metapruning}. Fundamentally there are three paradigms \cite{vibashan2022meta} in meta-learning; meta-optimizer, meta-representation and meta-objective. Meta-optimizer is the optimizer used to learn the optimization in the outer loop of meta-learning \cite{houthooft2018evolved}. Meta-representation aims to learn and update the meta-knowledge \cite{finn2017model}. Lastly, Meta-objective is the final achieved task after completing training \cite{li2019feature}. 

\textbf{Learning to prune filters}: Reinforcement Learning (RL) algorithms have been used to generate network architecture descriptions using Recurrent Neural Networks trained via policy gradient method \cite{zoph2016neural}. The same has also been implemented using Q-learning \cite{silver2016mastering}. Try-and-learn algorithm \cite{huang2018learning} uses RL to compute the reward of each filter and these rewards are then used to rank filters. It aggressively prunes filters in the baseline network while maintaining performance at a desired level. The model computes a reward as a product of the accuracy and efficiency term, and then uses REINFORCE \cite{williams1992simple} to estimate the gradients. The gradients can then be used to compute the loss and train the network, which learns to prune filters. The reward function makes it possible to control the trade-off between network performance and scale without human intervention. The try-and-learn algorithm automatically discovers redundant filters and removes them by repeating the process for every layer.

\textbf{Neural architecture searching}: Many methods have been proposed to search for optimal network structures from possible neural architectures \cite{zoph2016neural, tancik2021learned, wu2019fbnet}. There are primarily five methods used to search for an optimized network; reinforcement learning \cite{baker2016designing, zoph2016neural}, genetic algorithms \cite{xie2017genetic, real2017large}, gradient-based approaches \cite{wu2019fbnet}, parameter sharing \cite{cai2018proxylessnas} and weights prediction \cite{brock2017smash}.  \cite{zoph2016neural} uses RL to optimize the networks generated from model descriptions given by a recurrent network. MetaQNN \cite{baker2016designing} uses RL with a greedy exploration strategy to generate high-performing CNN architectures. Genetic algorithms are used to solve search and optimization problems using bio-inspired operators \cite{xie2017genetic}. \cite{real2017large} uses genetic algorithms to discover neural network architectures, minimizing the role of humans in the design. Gradient-based learning allows a network to efficiently optimize new instances of a task \cite{tancik2021learned}. FBNets (Facebook-Berkeley-Nets) \cite{wu2019fbnet} uses a gradient-based method to optimize CNNs created by a differentiable neural architecture search framework. 


\section{Rewarded meta-pruning}
\label{sec:proposed_method}

The Rewarded meta-pruning algorithm proposes using a reward coefficient to control the trade-off between the accuracy and efficiency of each model instead of finding the model with the highest accuracy within a preset range of FLOPs. The goal is to maximize the reward, which is directly proportional to accuracy and inversely proportional to the efficiency of the model. This method is implemented in three phases: training, searching, and retraining as shown in Algorithm \ref{alg:rewarded_meta_pruning}.

\subsection{Training}

Most popular CNNs \cite{he2016deep, howard2017mobilenets, sandler2018mobilenetv2} mainly use three types of layers; convolution block, bottlenecks, and linear layers. The channel scale represents the size of each layer. For the initial convolutional block and each type of bottleneck, the batch normalization with the sizes ranging from 10\% to 100\% the size of the original architecture, equally distributed between 31 initial weights. 


\begin{figure}[ht]
    \centering
    \includegraphics[scale=0.32]{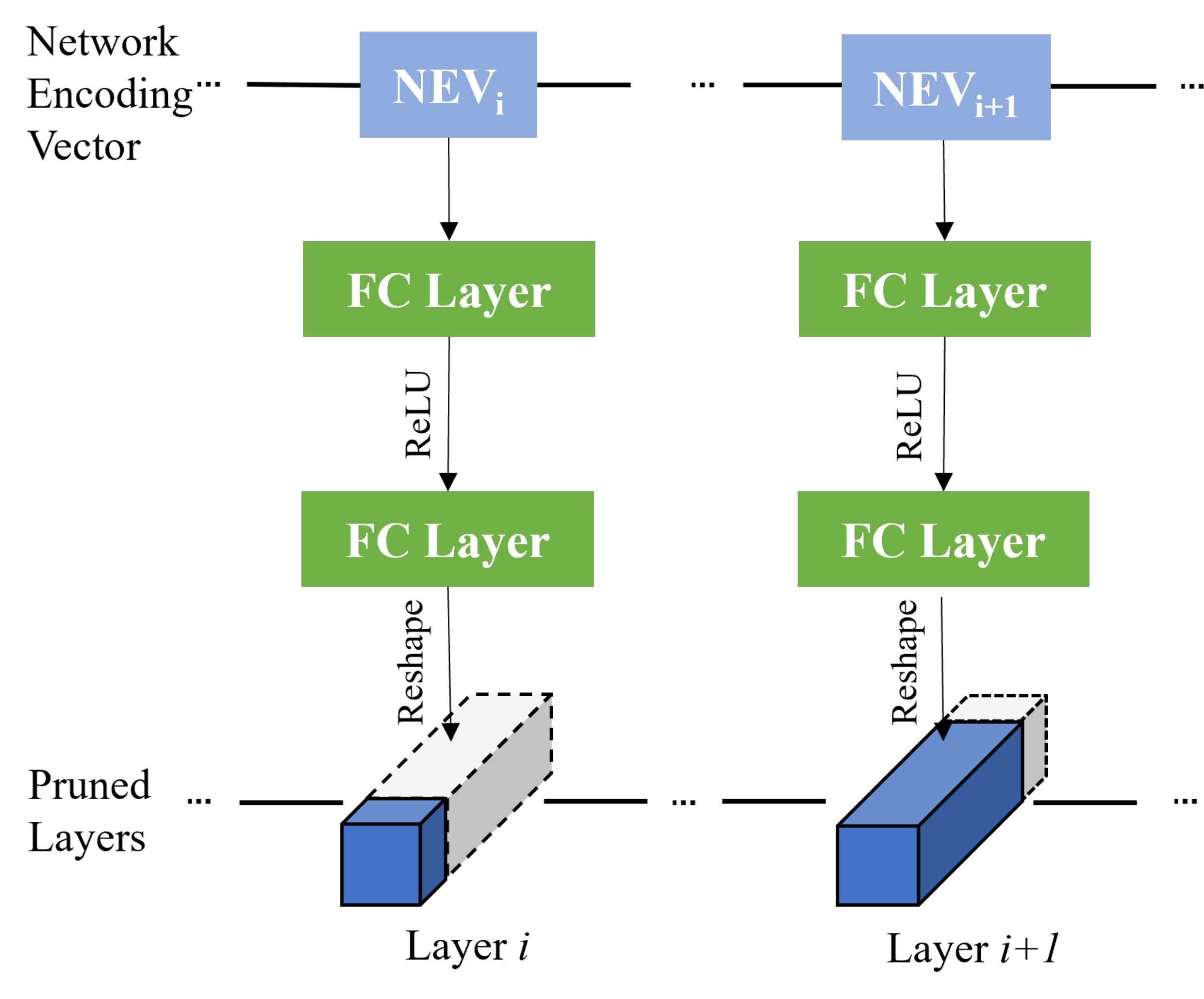}
    \caption{Stochastic training method}
    \label{fig:training}
\end{figure}


Each model is defined by a NEV, which is a list of random numbers defining the scale by which each layer is pruned from the 31 scales and their corresponding Batch Normalisation weights. It also creates linear layers at the scale defined by the NEV. The NEV is passed into two Fully Connected (FC) layers which generate weight matrix as shown in Figure \ref{fig:training}. The weights of these layers are trained by gradients of the generated weights calculated w.r.t the weights of the FC layers. The weights generated for each combination of output filter sizes are uniquely mapped because the function which converts a NEV to weights is an injective function. Thus, for each epoch, a random NEV creates a new model with weights that map one-to-one to the sample space of NEVs. These generated weights are reshaped. Each value $C_i$ in the NEV $C$ corresponds to the output channels of the layer $i$.

Once a model is created, it is trained for each batch of the train data with these initial weights, and cross-entropy loss is computed to update the weights. Each model is essentially a slice of a complete model, where the slice is defined by the NEV. Validation data is not used to validate the model, instead only to measure the progress. 


\begin{figure*}
    \centering
    \includegraphics[scale=0.59]{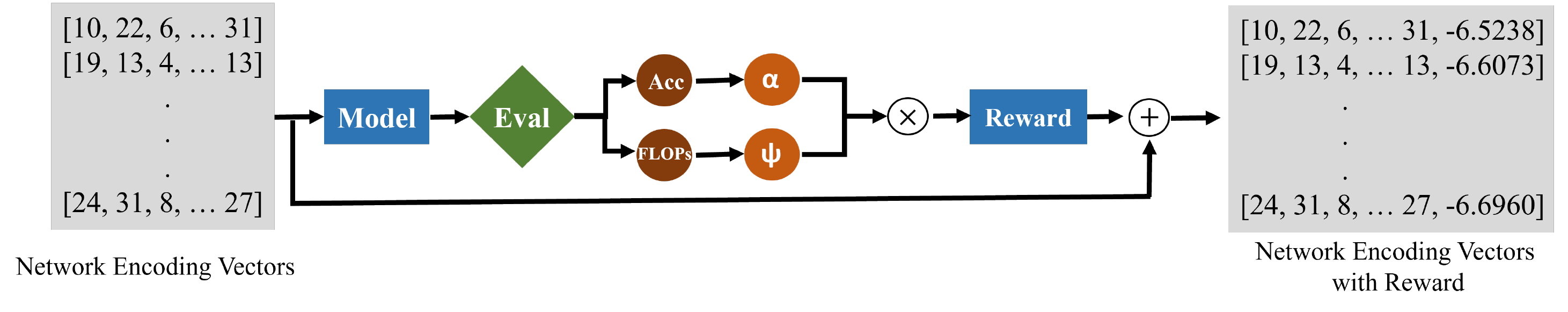}
    \caption{Computing reward for network encoding vectors.}
    \label{fig:nev_reward}
\end{figure*}

\subsection{Searching}

The models created from the NEV candidates use the weights of the trained model to create a pruned model. Each NEV is thus converted to a pruned model and reward is calculated for them. Evolutionary search \cite{mallipeddi2011differential} is then used to find the best NEV, thereby finding the optimal pruned model. The initial weights are the trained weights, but the final model will be trained from scratch to remove bias in the pre-trained model. 

\subsubsection{Creating genes} 
Random candidates are generated to seed the evolutionary search. Each gene is created as a list of sizes corresponding to the model with values representing the weights from the dictionary. However, since the metrics of the models created by these NEVs are also random, arbitrary hyperparameters are used to control the final model. A gene is considered valid if the FLOPs of the model created between $max\_FLOPs$ and $min\_FLOPs$. The FLOPs of each gene are stored as the last element of the NEV to reduce overall computation time. This is later replaced by the reward.

\subsubsection{Reward and selection of NEVs} 
The candidates are ranked after each epoch according to the reward, computed as a product of accuracy and efficiency coefficients given by Equations \eqref{eq:alpha} and \eqref{eq:psi} for each NEV as shown in Figure \ref{fig:nev_reward}. The reward is computed as the following formula:

\begin{equation}
    R(G_i) = \alpha (G_i,b_a) \times \psi (G_i,b_f ) .
\label{eq:reward}
\end{equation}

The accuracy and efficiency coefficients, denoted by $\alpha$  and $\psi$ , are defined as: 

\begin{equation}
    \alpha(G_i,b_a) = \left(\frac{b_a}{b_a-A(G_i)}\right)^2 ,
\label{eq:alpha}
\end{equation}

\begin{equation}
    \psi(G_i,b_f) = \log\left(\frac{b_f}{F(G_i)}\right),
\label{eq:psi}
\end{equation}
where $G_i$ denotes a gene of index $i$ from all candidate genes $G$, and $A$ and $F$ represent functions that return the accuracy and FLOPs of the model created using the gene passed into them.
The accuracy coefficient $\alpha$ increases exponentially with an increase in model accuracy, but as it approaches the base accuracy, the value tends to infinity. Since the model is not fine-tuned, the accuracy does not get close enough to the accuracy of the base model, $b_a$, for the reward to reach high levels. If knowledge distillation is used to increase the accuracy of the new model, the base accuracy would be the accuracy of the new model, thereby eliminating any negative effect from the symmetric nature of equation \ref{eq:alpha}. On the other hand, the efficiency coefficient $\psi$ linearly decreases with increasing FLOPs but is limited by the FLOPs of the original model $b_f$. 
Since prune rates are inversely proportional to FLOPs, a lower efficiency coefficient corresponds to a higher prune rate for the most part. 
The reward function is directly proportional to accuracy and prune rate.

The accuracy coefficient is directly proportional to the reward but is moderated by the efficiency coefficient. This creates a balance between them so that high accuracy is not achieved at the cost of low prune rates in the final model. Once a reward is computed, it is stored as the last element of the NEV, which is later used to rank each NEV. The Top-50 NEVs from every epoch is stored, then the 10 best of them are mutated and crossed over to get the candidate genes for the next epoch.

\subsubsection{Mutation and crossover} 
The best candidates from each epoch are mutated and crossed over to create candidates for the next epoch. Mutation involves changing a few elements in a gene to create a new gene. There is a 10\% chance for each element in a gene to be changed to a random valid element. Crossover is the combining of two random genes to create a new gene. An element is randomly picked from one of the two chosen genes for each index. Channel configurations in a local region of the configuration space tend to have similar metrics \cite{li2022revisiting}, so the new candidates also have similar accuracy and FLOPs. This makes the reward of at least the best candidate tend not to decrease. If the reward has not increased in too many epochs, the genetic search is stuck in local minima. However, since evolutionary search is a high-dimensional non-convex search, the critical points with errors much larger than that of the global minima are likely to be saddle points \cite{dauphin2014identifying}. In other words, the found local minima are likely to be close enough to the global minima, so the search can be terminated. It is not unlikely that mutating and crossing over more genes could find genes, but increasing the rates of mutations and crossover would affect the integrity of the evolutionary search. If unable to create enough new candidates from the two evolutionary operators, the remaining are created using random genes.


\begin{algorithm}
\caption{Algorithm of Rewarded meta-pruning}
\label{alg:rewarded_meta_pruning}
\begin{algorithmic}
\Statex \textbf{Hyperparameters: } $max\_training$: Number of training epochs, $max\_iter$: Number of searching epochs, $max\_tuning$: Number of finetuning epochs
\Statex \textbf{Input: } $dataset$: training images that can be split into $batches$, $r_i$: Random integer indexed at $i$, $w_i$: Random weights indexed at $i$, $\nabla$: Gradient of loss of given model
\Statex \textbf{Functions: } $norm(nev)$ converts $nev$ to weights, $FC(weights)$ creates model using $weights$, $f(model, data)$ trains model using given data, $reward(nev)$ computes reward of model created using $nev$, $mutation$ and $crossover$ performs evolutionary operations on list of $nev$s
\Statex \textbf{Output: } $x$: Pruned and trained model
\For{$i$ = 0,1,..., $max\_training$}
    \For{each $batch$ in $dataset$}
        \State $nev$ = [$r_1$,$r_2$,...,$r_n$]
        \State \{$w_1$,$w_2$, ... ,$w_n$\} = norm($nev$)
        \State $x$ = FC(\{$w_1$,$w_2$, ... ,$w_n$\})
        \State $L$ = f($x$,$batch$)
        \State $x$ += $\nabla L$
    \EndFor
\EndFor
\State $candidate$ = List of $n$ random $nev$s
\For{$i$ = 0,1,..., $max\_iter$}
    \State $rewards$ = [$r_1$,$r_2$,...,$r_n$]
    \For{$j$ = 0,1,..., $n$}
        \State $r_j$ = reward($nev_j$)
    \EndFor
    \State sort $candidate$ in descending order of $rewards$
    \State $mutated$ = mutation($candidate[:10]$)
    \State $crossed\_over$ = crossover($candidate[:10]$)
    \State $candidate$ = $mutated$ + $crossed\_over$
\EndFor
\State \{$w_1$,$w_2$, ... ,$w_n$\} = norm($candidate[0]$)
\State $x$ = FC(\{$w_1$,$w_2$, ... ,$w_n$\})
\For{$i$ = 0,1,..., $max\_tuning$}
    \State x += $\nabla$f($x$,$dataset$)
\EndFor
\end{algorithmic}
\end{algorithm}


\subsection{Retraining}

Once the evolutionary search has been completed, the best gene is selected as the first gene in the list of candidates. This is the gene with the highest reward as found after multiple epochs of genetic searching. The best NEV is converted to a model and trained from scratch. All pruning algorithms train a pruned model for a few epochs to regain the lost accuracy in a process called fine tuning \cite{han2015learning}. In the Rewarded meta-pruning algorithm, model is created from an NEV instead of using the NEV to prune, and then trained from scratch. Hence the accuracy saturates at a higher epoch during finetuning.


\section{Experimental Results}

In this section, we demonstrate the superiority of the Rewarded meta-pruning method. We first describe the experimental settings to reproduce the experiments. Then we compare the results obtained with other methods pruning three major networks. Lastly, we perform an ablation study to understand the effectiveness of the proposed method.

\subsection{Experimental setting}

ResNet-50 \cite{he2016deep} network is trained for 32 epochs, while MobileNetV1 \cite{howard2017mobilenets} and MobileNetV2 \cite{sandler2018mobilenetv2} are trained for 64 epochs. ResNet and MobileNetV2 retrained after searching for 400 epochs, but MobileNetV1 only requires 320 epochs. Searching for the NEVs takes 20 epochs for all the networks. Each epoch searches 50 NEVs, searching through 1000 unique NEVs throughout the run. MobileNetV1 and MobileNetV2 both use the Lambda scheduler to decay the models by a $\gamma$ of 0.1 every epoch from an initial learning rate of 0.2. The scheduler for ResNet, however, decreases by a factor of 0.1 at epochs 80 and 160.

The experiments are conducted on three commonly used networks including ResNet-50, MobileNetV1, and MobileNetV2. The networks are trained using ImageNet \cite{deng2009imagenet} from scratch as described in the previous section. ImageNet consists of 1.2M training images and 50K validation images. It also consists of 100K test images, but since the labels of the test data are not released, validation data is used for testing. The validation data has not been used in any part of the training process except to compute the accuracy at every stage. As a natural consequence of using evolutionary computation, Rewarded meta-pruning is resource-heavy in computing the pruned networks. But this cost is balanced out by the efficiency and accuracy of the models generated. A network need not be created each time it is used because the knowledge distilled from it can be transferred and used in varying contexts.

\subsection{Evaluation protocol}

Four metrics are used for evaluating the pruning algorithms: parameter ratio, top-1 and top-5 errors and FLOPs. The parameter ratio is the ratio of the pruned model to the baseline model. It is computed as; 
\begin{equation}
    P^* = \frac{P_m}{P_b} \times 100\% ,
\label{eq:parameter }
\end{equation}
where $P_m$ and $P_b$ are the numbers of weights in the pruned and the baseline model respectively. Accuracy is the percentage of validation data identified correctly compared to the whole dataset. Top-1 error is the inverse of accuracy, i.e., the proportion of images where the predicted labels of the highest probability are wrong. Top-5 error is the proportion of images where the correct label is not present in the five highest probabilities of predicted labels. FLOPs is a measure of the number of Floating-point operations computed per second.

\subsection{Performance on ResNet-50}

ResNet-50 is a CNN with a depth of 50 layers. It was created to solve degradation in the model as deeper layers are stacked \cite{he2016deep}. ResNet uses skip connections to identify mapping. This adds the features with their original parameters before passing them into the next layer. Identity mapping followed by linear projection is used to expand channels of the features to make it possible to be added with the original parameters.

\begin{table}[ht]
\centering
\resizebox{\columnwidth}{!}{%

\begin{tabular}{ccccc}
\toprule
\textbf{Method} & \textbf{Top-1 Error} & \textbf{Top-5 Error} & \textbf{FLOPs} \\
\midrule
Baseline \cite{li2022revisiting} & 23.40\% & - & 4110M \\
GAL-0.5 \cite{lin2019towards} & 28.05\% & 9.06\% & 2341M \\
SSS \cite{huang2018data} & 28.18\% & 9.21\% & 2341M \\
HRank \cite{lin2020hrank} & 25.02\% & 7.67\% & 2311M \\
Random Pruning \cite{li2022revisiting} & 24.87\% & 7.48\% & 2013M \\
AutoPruner \cite{luo2020autopruner} & 25.24\% & 7.85\% & 2005M \\
Adapt-DCP \cite{liu2021discrimination} & 24.85\% & 7.70\% & 1955M \\
MetaPruning \cite{liu2019metapruning} & 24.60\% & - & 2005M \\
\textbf{Rewarded meta-pruning} & \textbf{24.24\%} & \textbf{7.35\%} & \textbf{1950M} \\
\bottomrule
\end{tabular}
}
\caption{ Benchmarking state-of-the-art channel pruning methods with ResNet-50 }
\vspace{-0.01in}
\label{tab:table_Res50}
\end{table}

Table \ref{tab:table_Res50} shows the results of ResNet-50 trained using ImageNet-2012 after pruning with Rewarded meta-pruning and other competing methods. It can be inferred that this method has a lower error rate than every method, and this is achieved while keeping the FLOPs relatively low. The FLOPs, as compared to the baseline network \cite{li2022revisiting}, reduced by 52.55\%, but the error has only increased by 0.84\%. When compared with standard random pruning, for a similar reduction in FLOPs, there is a 0.63\% lower error. MetaPruning \cite{liu2019metapruning} method shows a 0.36\% higher error while still using 1.34\% higher FLOPs as compared to the baseline model. Adapt-DCP \cite{liu2021discrimination} has the closest reduction in FLOPs as compared to the baseline, but this method has a 0.61\% lower error. SSS \cite{huang2018data} and HRank \cite{lin2020hrank} methods have very similar prune rates to the Rewarded meta-pruning, but have higher FLOPs by around 9\%, and higher error by 3.94\% and 0.78\% respectively. 

\subsection{Performance on MobileNetV2}

MobileNetV2 contains depth-wise and point-wise convolution. It has an inverted residual with a linear bottleneck which takes a low dimensional compressed representation as input and expands it to a higher dimension, then filters them with light-weight depthwise convolutions like in MobileNetV1 \cite{sandler2018mobilenetv2}. MobileNetV2 is an efficient network with a relatively low error. Thus, demonstration on MobileNetV2 is an effective way to show the performance of the pruning algorithm.

\begin{table}[ht]
\centering
\resizebox{\columnwidth}{!}{%

\begin{tabular}{ccccc}
\toprule
\textbf{Method} & \textbf{Top-1 Error} & \textbf{Top-5 Error} & \textbf{FLOPs} \\
\midrule
Baseline \cite{li2022revisiting} & 28.12\% & 9.71\% & 314M \\
0.75 MobileNetV2 \cite{li2022revisiting} & 30.20\% & - & 220M \\
Random Pruning \cite{elkerdawy2022fire} & 29.10\% & - & 223M \\
AMC \cite{he2018amc} & 29.20\% & - & 220M \\
MetaPruning \cite{liu2019metapruning} & 28.80\% & - & 227M \\
Greedy Selection \cite{ye2020good} & 28.80\% & - & 201M \\
Adapt-DCP \cite{liu2021discrimination} & 28.55\% & - & 216M \\
\textbf{Rewarded meta-pruning} & \textbf{28.51\%} & \textbf{10.65\%} & \textbf{199M} \\
\bottomrule
\end{tabular}
}
\caption{ Benchmarking state-of-the-art channel pruning methods with MobileNetV2 }
\vspace{-0.01in}
\label{tab:table_MNV2}
\end{table}

Table \ref{tab:table_MNV2} compares the performance of the Rewarded meta-pruning method with the state-of-the-art methods. The Rewarded meta-pruning method has a lower FLOPs than any other methods while showing only 0.39\% higher error than the baseline. 0.75 MobileNetV2, which is MobileNetV2 of width 25\% lower than the original, has 1.69\% higher error than this method. When compared to random pruning, which is the baseline for all pruning methods \cite{blalock2020state}, this method has a 0.59\% lower error. MetaPruning \cite{liu2019metapruning} has a 0.29\% higher error despite having a 7.64\% higher FLOPs. AMC \cite{he2018amc} and Adapt-DCP \cite{liu2021discrimination} have 0.69\% and 0.04\% higher error and FLOPs. Rewarded meta-pruning also outperforms Greedy selection \cite{ye2020good} by 0.29\% in spite of an almost similar amount of FLOPs. 

\subsection{Performance on MobileNetV1}

MobileNetV1 has a streamlined architecture that builds lightweight deep networks using depth-wise separable convolutions. All layers use Batch Normalisation and ReLu, except the fully connected layer which is followed by a softmax layer for classification \cite{howard2017mobilenets}.

\begin{table}[ht]
\centering
\resizebox{\columnwidth}{!}{%

\begin{tabular}{ccccc}
\toprule
\textbf{Method} & \textbf{Top-1 Error} & \textbf{Top-5 Error} & \textbf{FLOPs} \\
\midrule
Baseline \cite{howard2017mobilenets} & 29.40\% & - & 569M \\
0.75 MobileNet-224 \cite{howard2017mobilenets} & 31.60\% & - & 325M \\
FTWT (r=1.0) \cite{elkerdawy2022fire} & 30.34\% & - & 335M \\
MetaPruning \cite{liu2019metapruning} & \textbf{29.10}\% & - & 324M \\
\textbf{Rewarded meta-pruning} & 29.60\% & \textbf{9.65\%} & \textbf{295M} \\
\bottomrule
\end{tabular}
}
\caption{ Benchmarking state-of-the-art channel pruning methods with MobileNetV1 }
\vspace{-0.01in}
\label{tab:table_MNV1}
\end{table}

In Table \ref{tab:table_MNV1}, we compare the Rewarded meta-pruning method with other competing pruning techniques to prune a MobileNetV1 model. 
It has almost regained the accuracy of the baseline network \cite{howard2017mobilenets}, with 0.2\% lower accuracy and 48.15\% lower FLOPs. 
This method clearly achieves superior results when compared to Fire-Together-Wire-Together \cite{elkerdawy2022fire} pruning method, with a 0.94\% lower error and 7.03\% lower FLOPs. 
It outperforms 0.75 MobileNet-224 \cite{howard2017mobilenets}, which is MobileNetV1 with 25\% lower width, by 2\% while using 5.27\% lower FLOPs. 
While MetaPruning \cite{liu2019metapruning} shows lower error than even the baseline network, our method has a lower FLOPs. However, the size of the pruned network is still 9.83\% larger, when compared to the proposed method. This could be due to the lack of shortcut connection in MobileNetV1 in spite of being a smaller network, leading to a large number of fully-connected layers. In terms of performance achieved for ever resource, Rewarded meta-pruning edges out MetaPruning. It is fair to assume that the Rewarded meta-pruning method could have better results with more advanced reward functions. This will be validated by further research on the robustness of various hyperparemeters.

\subsection{Discussion}

From the results, it is clear that the proposed method performs best under the right reward functions. The reward function in this case is directly proportional to the accuracy and inversely proportional to FLOPs. 

As the accuracy of the pruned model approaches the baseline accuracy, the reward increases exponentially. In MetaPruning \cite{liu2019metapruning}, the reward is directly proportional to accuracy, whereas in this method, the reward is proportional to the square of the accuracy of the model. This by itself would not increase the accuracy of the final model, but chasing a higher reward that is only dependent on accuracy could lead to the final model having high FLOPs. This can be seen in MetaPruning, where the pruned model has a tendency to show FLOPs as high as the preset maximum FLOPs would allow. The reward is controlled by the efficiency coefficient to prevent this.

The reward decreases with increasing FLOPs because the efficiency coefficient is inversely proportional to the FLOPs. However, if the proportionality is too high, the reward would be throttled. Hence it is controlled by the linearly decreasing efficiency coefficient.

\begin{figure}[ht]
    \centering
    \includegraphics[scale=0.27]{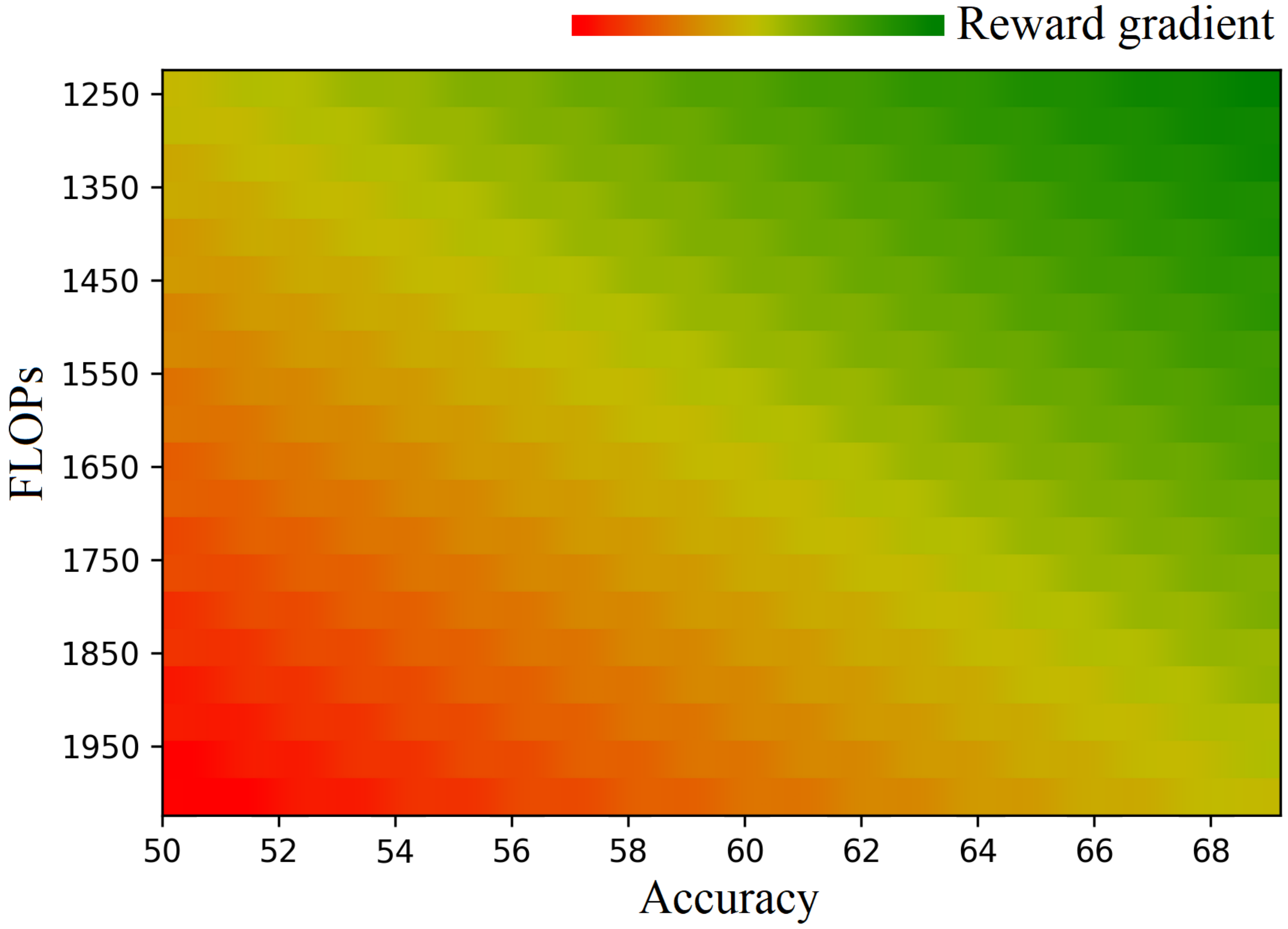}
    \caption{Reward at varying Accuracies and FLOPs.}
    \label{fig:heatmap_graph}
\end{figure}

From the definition of the reward function in Equation \eqref{eq:reward}, it is clear that high accuracy alone is not enough for a model to be selected. As the accuracy increases, the probability of the model being selected increases, but only so long as the FLOPs of the model are also low enough. This can be inferred from how the distribution of $Green$ increases as we move towards the right as shown in Figure \ref{fig:heatmap_graph}. The reward increases as the FLOPs decrease, but it is not deemed acceptable until accuracy is high enough. This can be inferred from how the distribution of $Red$ increases as the FLOPs decrease as inferred from Figure \ref{fig:heatmap_graph}. Thus by definition, the Rewarded meta-pruning method leans more toward accuracy. 


\begin{figure*}
    \centering
    \includegraphics[scale=0.59]{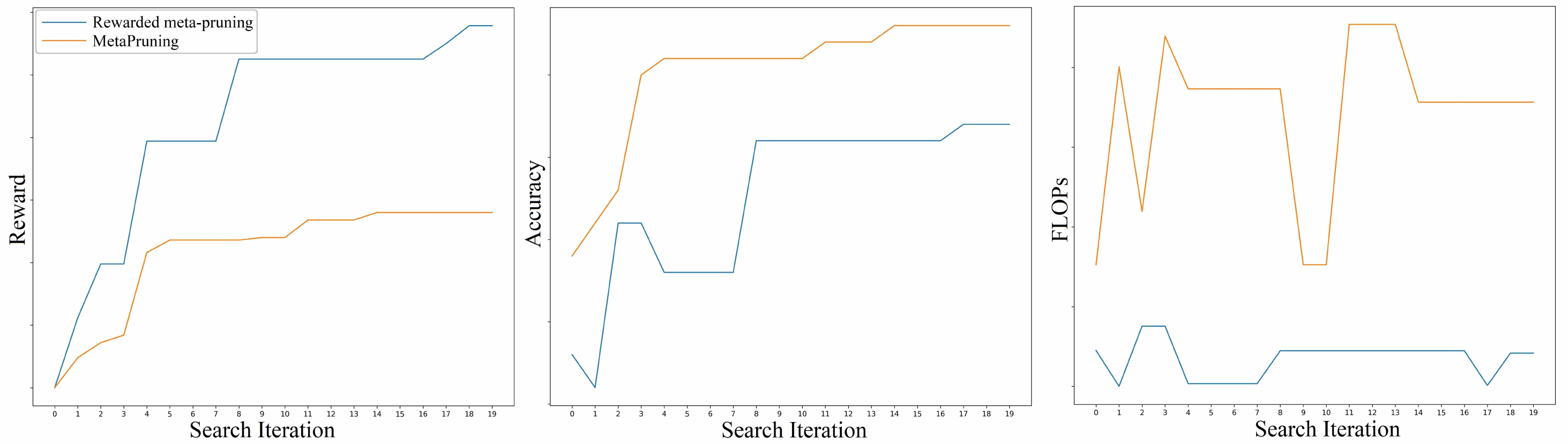}
    \caption{Rewards, Accuracy and FLOPs of the best models after each iteration of searching.}
    \label{fig:slopes}
\end{figure*}

When compared to MetaPruning, the reward of the Rewarded meta-pruning method increases higher with each iteration of searching. This is because the reward of MetaPruning is directly proportional to accuracy while the reward of Rewarded meta-pruning is proportional to the square of accuracy. This can be observed in Figure \ref{fig:slopes} which shows the rewards, accuracies, and FLOPs of the best model after each iteration of searching. The accuracy and FLOPs, in the beginning, are chosen randomly, but that cannot be changed without tampering with the fundamental evolutionary searching. But it can be observed that the FLOPs of the best model in MetaPruning tend to increase for the most part whereas, in the case of Rewarded meta-pruning method, it tends to stay low. Accuracy increases in both cases, but MetaPruning saturates earlier than Rewarded meta-pruning. It can also be inferred that if the two models started with the same batch of randomly initialized models, the Rewarded meta-pruning method will have a higher accuracy and a lower FLOPs because the slopes of the best accuracy and FLOPs of the best models are higher and lower respectively than that of MetaPruning.

\begin{figure}[ht]
    \centering
    \includegraphics[scale=0.6]{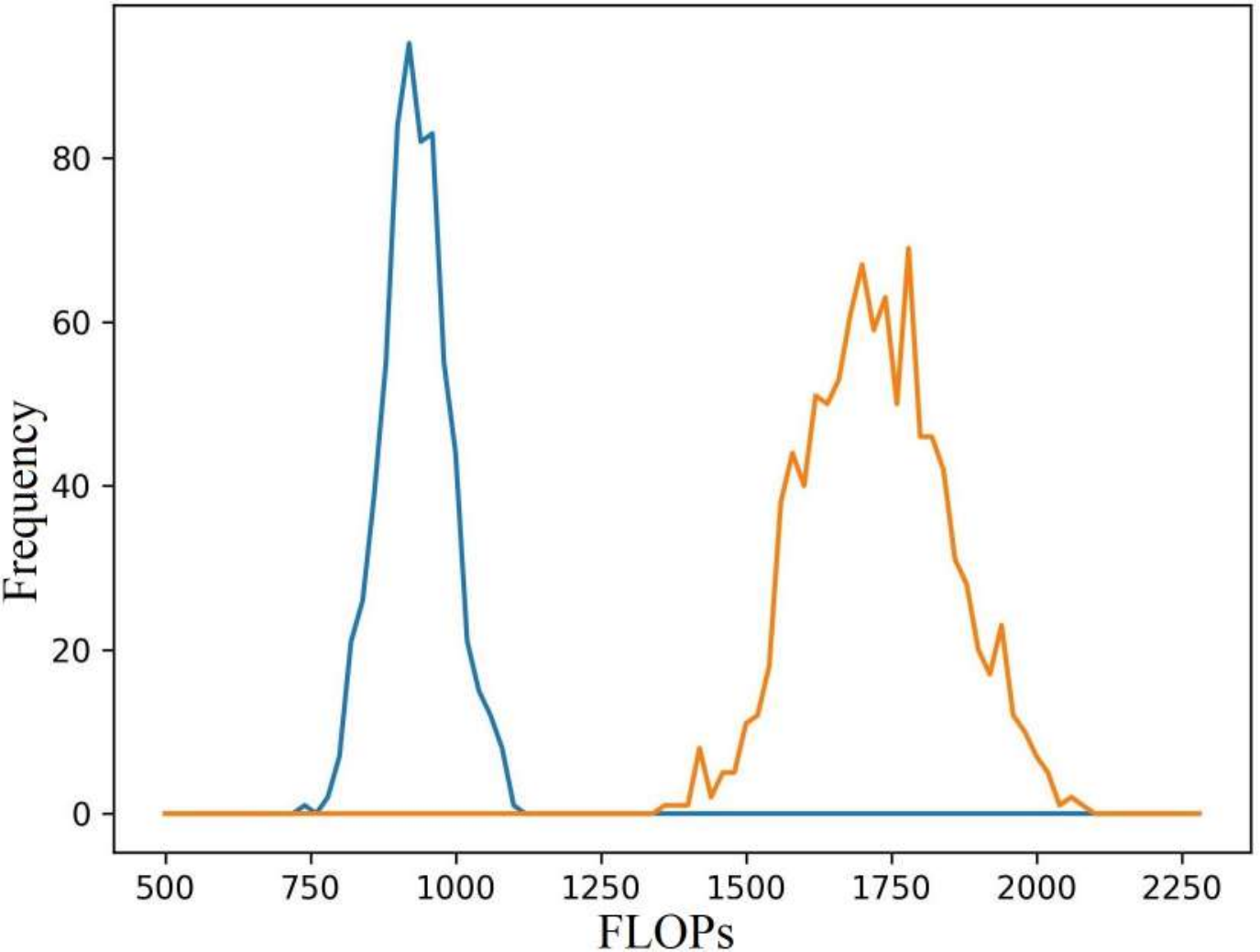}
    \caption{Distribution of FLOPs of 1000 randomly generated NEVs with varying ranges of FLOPs.}
    \label{fig:var_flops}
\end{figure}

However, the way NEVs are chosen, both randomly and after mutation or crossover, means the FLOPs of the pruned model approximately form a bell-curve between 1350 and 2100. This can be changed by controlling the generation of random filter sizes in the NEVs as shown in Figure \ref{fig:var_flops}. 

The robustness of the hyperparameters used by Rewarded meta-pruning has already been explored by He \etal \cite{he2020learning}. As the reward function is tweaked to add different hyperparameters, various metrics of the final model can be controlled. There are various other coefficients that could be used in the reward. The prune rate of the pruned model can be set to be inversely proportional to the reward, as low prune rates automatically lead to lower FLOPs. The FLOPs is also directly proportional to hardware latency, which is the runtime of networks \cite{dong2018dpp}. But this is dependent on hardware, and different hardware have different latencies. Other metrics such as energy consumption have also been used for pruning. NetAdapt \cite{yang2018netadapt} uses energy consumption as a metric to measure the complexity of the network at every stage and prunes the network further while maintaining accuracy.


\section{Conclusion}

In this work, we have presented the following: 
1) Implemented better reward function to meta-learn parameters for pruning and allow better control over various parameters of the pruned model. 
2) The Rewarded meta-pruning method has been shown to be superior to other state-of-the-art methods, with higher accuracy and lower FLOPs than traditional channel pruning methods. 
3) The reward function can be optimized using other metrics to maximize the reward. 
4) ResNet-50, MobileNetV1 and MobileNetV2 are pruned effectively. 


\section{Acknowledgement}

This work was supported by the National Research Foundation of Korea (NRF) grant funded by the Korean Government (MSIT) (No.2021R1C1C1012590), (No. NRF-2022R1A4A1023248) and the Information Technology Research Center (ITRC) support program supervised by the Institute of Information Communications \& Technology Planning \& Evaluation (IITP) grant funded by the Korean Government (MSIT) (IITP-2022-2020-0-01808).


{\small
\bibliographystyle{ieee_fullname}
\bibliography{egbib}
}


\end{document}